\documentclass[runningheads]{llncs}
\usepackage{amsfonts} 
\usepackage[T1]{fontenc}
\usepackage{graphicx}
\usepackage{blindtext}
\usepackage{tabularx}
\usepackage{url}
\begin{document}
\title{Classification of Visualization Types and Perspectives in Patents}
%
%
\author{Junaid Ahmed Ghauri\inst{1}\orcidID{0000-0001-9248-5444} \and
Eric M\"uller-Budack\inst{1,2}\orcidID{0000-0002-6802-1241} \and
Ralph Ewerth\inst{1,2}\orcidID{0000-0003-0918-6297}}
\authorrunning{Ghauri et al.}
%
\institute{TIB -- Leibniz Information Centre for Science and Technology, Hannover, Germany \and
L3S Research Center, Leibniz University, Hannover, Germany\\
}
\maketitle              
\begin{abstract}
Due to the swift growth of patent applications each year, information and multimedia retrieval approaches that facilitate patent exploration and retrieval are of utmost importance.
Different types of visualizations~(e.g., graphs, technical drawings) and perspectives~(e.g., side view, perspective) are used to visualize details of innovations 
in patents. 
The classification of these images enables a more efficient search and allows for further analysis. 
%
So far, datasets for image type classification miss some important visualization types for patents. Furthermore, related work 
does not make use of recent deep learning approaches including transformers. 
%
In this paper, we 
adopt state-of-the-art deep learning methods for the classification of visualization types 
and perspectives in patent images.
We extend the \textit{CLEF-IP} dataset for image type classification in patents to ten classes and provide manual ground truth annotations. In addition, we derive a set of hierarchical classes from a dataset that provides weakly-labeled data for image perspectives. 
Experimental results have demonstrated the feasibility of the proposed approaches.
%
Source code, models, and dataset will be made publicly available\footnote{\url{https://github.com/TIBHannover/PatentImageClassification}}.

\end{abstract}
%
%
%

\section{Introduction}

Patents are legal documents that represent intellectual property to exclude others from making, using, or selling inventions.
The number of patent applications submitted to patent organizations like \textit{WIPO}~(World Intellectual Property Organization), \textit{EPO}~(European Patent Office), and \textit{USPTO}~(United States Patent and Trademark Office) is rapidly rising. 
For example, the \textit{WIPO} received more than three million patent applications in 2021\footnote{\url{https://www.wipo.int/en/ipfactsandfigures/patents}}. 
Details of the inventions proposed in patents are typically presented using text and images~\cite{krestel_survey_paper}.
Different visualization types are used to efficiently convey information, e.g., block diagrams, graphs, and technical drawings~\cite {shuo_pat_cnn_paper}. In some cases, technical drawings are illustrated in more than one perspective~(e.g., top or front view) to depict details~\cite{xinwei_visual_captions_paper}. 
%
Novel information and multimedia retrieval methods are necessary to facilitate the search and exploration of patents, e.g., to allow human assessors to find relevant patents, evaluate the novelty, 
or find possible plagiarisms~\cite{krestel_survey_paper,Hideo_survey_paper}.

%
According to a recent survey paper on patent analysis~\cite{krestel_survey_paper}, there has been a lot of progress in tasks like 
patent retrieval~\cite{Pustu_multimodal_pat_analysis_paper,Liping_pat_image_retrieval_paper,Stefanos_pat_image_retrieval_paper,Florina_pat_image_retrieval_paper} and
patent image classification~\cite{shuo_pat_cnn_paper,xinwei_visual_captions_paper,michal_deeppat_paper}
due to the advancements in deep learning. 
%
%
We mainly focus on image classification since visualizations contain important information about patents~\cite{krestel_survey_paper,allan_survey_paper,Hideo_survey_paper}. 
However, patents can depict various visualization types that require specific information extraction techniques, e.g. for tables~\cite{PaliwalDRSV19,Nazir_table_from_scanned_documents,Gralinski_information_extraction,Schreiber_icdar_table_from_document} or structured diagrams~\cite{DBLP:conf/eccv/KembhaviSKSHF16,Lee_Pix2Struct,wang_computer_science_diagrams,XinHu_structure_diagram_object_detection}. 
%
%
So far, there have been some approaches for image type classification in scientific documents~\cite{DBLP:conf/ecir/MorrisME20,DBLP:conf/icdar/JobinMJ19} but the application domain and images differ in terms of style, structure, etc. compared to patent images.
%
%
For patents specifically, Jiang et al.~\cite{shuo_pat_cnn_paper} suggested a deep learning model for image type classification and applied it to 
the \textit{CLEF-IP 2011 dataset}~\cite{Florina_pat_image_retrieval_paper}. 
However, existing datasets~\cite{Florina_pat_image_retrieval_paper,shuo_pat_cnn_paper,michal_deeppat_paper} on image type classification contain different classes that miss some important types used in patents. 
%
%
The image perspective is another important aspect 
since it helps to analyze technical aspects of the same drawing from different viewing angles.
To the best of our knowledge, there is only one approach for the classification of image perspectives~\cite{xinwei_visual_captions_paper} which only considers textual information from captions to determine the perspective of the associated image. 
%
Overall, works for visualization type and perspective classification have not leveraged recent deep learning approaches~\cite{efficientnet_mingxing,resnext_saining,regnet_ilija,alec_clip_paper} that have achieved tremendous progress in various image classification tasks. 
%
In particular, vision-language models such as CLIP~(Contrastive Language-Image Pretraining)~\cite{alec_clip_paper} have achieved impressive results in many downstream applications. 

%

In this paper, we address the aforementioned limitations. 
Our contributions can be summarized as follows: 
(1)~We present approaches that adopt state-of-the-art methods from computer vision~\cite{he_resnet_paper,resnext_saining,regnet_ilija,efficientnet_mingxing,alec_clip_paper} for patent image type and perspective classification. 
%
(2)~We extend the CLEF-IP dataset~\cite{Florina_pat_image_retrieval_paper,shuo_pat_cnn_paper} with the class of \textit{block circuit} as this can depict important technological innovations along with manual annotations to provide all ground-truth labels.
(3)~We extracted perspective class labels from 
dataset by Wei et al.~\cite{xinwei_visual_captions_paper} that uses textual descriptions for perspective detection. We also identified a class hierarchy with three levels of complexities and 2, 4, and 7 perspective classes, respectively.
(4)~We conduct an in-depth analysis of the proposed approaches on the datasets created for visualization type and perspective classification. Overall, we achieved promising results and provide strong baselines based on state-of-the-art approaches. 

The rest of the paper is organized as follows. 
Section~\ref{sec:methodology} describes the proposed approach and architecture for image type and perspective classification in patents. The experimental setup, dataset, and results are reported in Section~\ref{sec:experiments_results}. Section~\ref{sec:conclusion} concludes the paper and outlines potential future research directions.



\section{Image Type and Perspective Classification in Patents}
\label{sec:methodology}
\begin{figure}[t]
  \centering
  \includegraphics[width=0.91\linewidth]{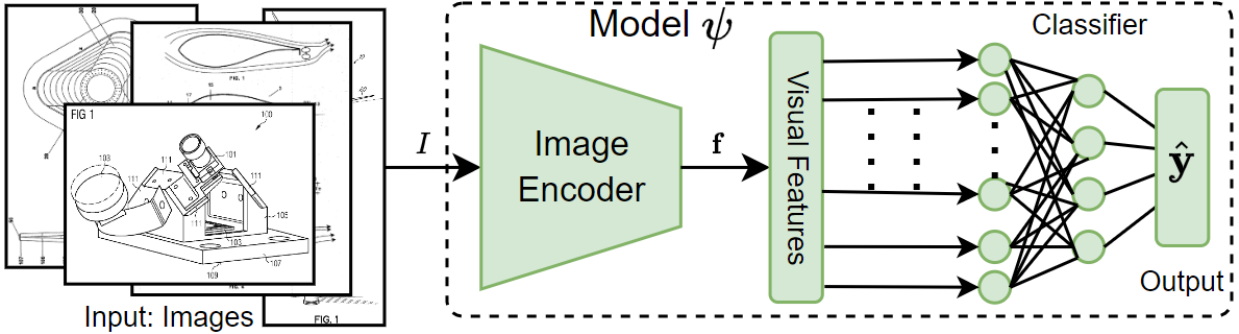}
  \caption{Pipeline for patent image classification including visual feature extraction and classification using a multi-layer perceptron~(MLP) for the respective classes.}
  \label{fig:uniModalPipeline}
\end{figure}

This section describes our proposed approaches for visualization type and perspective classification in patents. 
For these two individual tasks, the goal is to find models~$\psi(I)\,\to\ y$ that predict the correct class~$y$ for a given image~$I$. 
%
%
In general, we follow the pipeline illustrated in Figure~\ref{fig:uniModalPipeline}.
In the first step, we use deep learning models to extract visual features~$\mathbf{f}$ 
from patent images. Unlike related work, we also consider recent vision-language models such as \textit{CLIP}~\cite{alec_clip_paper}.
%
%
Second, we use a multilayer perceptron~(MLP) to predict the probabilities~$\hat{\mathbf{y}} = \langle \hat{y}_1, \hat{y}_2, \dots, \hat{y}_c\rangle$ for the $c$~classes of the given task. 
In the following, two approaches based on recent convolutional neural networks~(CNNs, Section~\ref{sec:cnn-image-classification}) and the vision-language transformer \textit{CLIP}~\cite{alec_clip_paper}~(Section~\ref{sec:clip-image-classification}) are described.
%
%

\subsection{Patent Classification using CNN Models}
\label{sec:cnn-image-classification}
We selected four CNNs~(notation in bold), i.e., 
\textit{\textbf{ResNet}-50}~\cite{he_resnet_paper}, 
\textit{\textbf{RegNet}Y-16GF}~\cite{regnet_ilija},
\textit{\textbf{EfficientNetV2}-M}~\cite{efficientnet_mingxing}, 
and \textit{\textbf{ResNeXt}-101(64$\times$4d)}~\cite{resnext_saining}, 
which have been proven to produce promising results in image classification tasks, as backbones to extract features~$\mathbf{f}$. 
%
%
We use the official implementations and set the number of neurons in the last fully-connected layer to the number of classes~$c$ for the given task.
During training, we finetune the weights of the \textit{entire network} using the cross-entropy loss between the one-hot encoded ground-truth vector~$\mathbf{y} = \langle y_1, y_2, \dots, y_c \rangle$ and the predicted probabilities~$\hat{\mathbf{y}}$. 
Finetuning the entire network including the image encoder typically provides better results than solely training a classifier since features specific to patent images can be learned~\cite{DBLP:conf/icml/HoulsbyGJMLGAG19,DBLP:conf/nips/LianZFW22}.

\subsection{Patent Classification using CLIP}
\label{sec:clip-image-classification}
To compare the CNN-based approach to recent vision-language models, we apply \textit{CLIP}~\cite{alec_clip_paper} that has achieved promising results in many downstream applications. 
We use the vision transformer~(ViT-B/32)~\cite{vit_models} of \textit{CLIP} to extract the visual features~$\mathbf{f}$ from the input images.
Since finetuning of transformers requires much more data~\cite{DBLP:conf/acl/LiL20,DBLP:conf/nips/LianZFW22}, we decided to freeze the weights of the image encoder during training. Instead, we use a multi-layer perceptron~(MLP) comprising three fully-connected layers with 256, 128, and 64 neurons to learn a feature representation for patent images. 
Finally, we apply another fully-connected layer to predict the probabilities~$\hat{\mathbf{y}}$ for $c$~classes. As for the CNN models~(Section~\ref{sec:cnn-image-classification}), we use the cross-entropy loss between the ground truth and predictions to optimize the MLP. In the remainder of this paper, this model is denoted as \textbf{\textit{CLIP+MLP}}.

%


\section{Experimental Setup and Results}\label{sec:experiments_results}
In this section, we present the experimental setup~(Section~\ref{sec:dataset_eval}) and results for visualization type~(Section~\ref{sec:results_image_type}) and perspective classification~(Section~\ref{sec:results_perspective_type}). 

\begin{table}[t] 
  \begin{center}
    \caption{Statistics for the extended CLEF-IP 2011~(left) and USPTO-PIP dataset with different granularities~(right) for two~($C_2$), four~($C_4$), and seven classes~($C_7$).} 
    \label{tab:datasetDistribution}
    \parbox{.4\linewidth}{
        \begin{tabular}{|l|c|c|c|c|}
         \multicolumn{2}{c}{}\\
           \hline
          \textbf{Image Type} & \textbf{Train} & \textbf{Val}& \textbf{Test}\\
          \hline
          Block Circuit & 450 & 50 & 100\\
          \hline
          Chemical & 5362 & 595 & 112\\
          \hline
          Drawing & 5009 & 556 & 274\\
          \hline
          Flowchart & 279 & 31 & 102\\
          \hline
          Genesequence & 5385 & 598 & 24\\
          \hline
          Graph & 1497 & 166 & 193\\
          \hline
          Maths & 5355 & 595 & 126\\
          \hline
          Program & 5016 & 557 & 26\\
          \hline
          Symbol & 1421 & 157 & 17\\
          \hline
          Table & 4952 & 550 & 66\\
          \hline
        \end{tabular}
    }
    \hfill
    \parbox{.56\linewidth}{
        \begin{tabular}{|l|c|c|c|c|c|c|}
        \hline
          \textbf{Perspective Type} & \textbf{Train} & \textbf{Val}& \textbf{Test}&$C_2$&$C_4$&$C_7$\\
          \hline
          \textbullet~Perspective View & 6140 & 150 & 150 &$\checkmark$ &$\checkmark$ &$\checkmark$ \\
          \hline
          \hline
          \textbullet~Non-Perspective & 18470 & 900 & 900
          &$\checkmark$ &$\times$&$\times$\\
          \qquad\textbullet~Left-Right & 4767 & 300 & 300
          &$\times$&$\checkmark$ &$\times$\\
          \qquad\qquad\textbullet~Left & 2407 & 150 & 150
          &$\times$&$\times$&$\checkmark$ \\
          \qquad\qquad\textbullet~Right & 2360 & 150 & 150
          &$\times$&$\times$&$\checkmark$ \\
          \hline
          \qquad\textbullet~Bottom-Top & 6060 & 300 & 300
          &$\times$&$\checkmark$ &$\times$\\
          \qquad\qquad\textbullet~Bottom & 2800 & 150 & 150 &$\times$&$\times$&$\checkmark$ \\
          \qquad\qquad\textbullet~Top & 3260 & 150 & 150 
          &$\times$&$\times$&$\checkmark$ \\
          \hline
          \qquad\textbullet~Front-Rear & 7643 & 300 & 300  
          &$\times$&$\checkmark$ &$\times$\\
          \qquad\qquad\textbullet~Front & 5184 & 150 & 150 &$\times$&$\times$&$\checkmark$ \\
          \qquad\qquad\textbullet~Rear & 2459 & 150 & 150 
          &$\times$&$\times$&$\checkmark$ \\
          \hline
        \end{tabular}
    }
  \end{center}
\end{table}

\subsection{Experimental Setup} 
\label{sec:dataset_eval}
%
%
In the following, we provide details on the datasets for visualization type and perspective classification, evaluation metrics, and implementation. 

\paragraph{Extended CLEF-IP 2011 Dataset:}
We use the 2011 benchmark dataset of CLEF-IP~\cite{Florina_pat_image_retrieval_paper} for visualization type classification. 
However, it does not cover \textit{block and circuit diagrams}, which are an important type of visualization frequently used in patents. We added this category as a tenth class and collected images by querying EPO's publication server~(\url{https://data.epo.org/publication-server/}). 
Finally, we manually annotated images depicting \textit{block and circuit diagrams}. 
The dataset statistics are provided in Table~\ref{tab:datasetDistribution}~(left). 

\paragraph{USPTO-PIP Dataset:}
%
For the perspective classification task, we used the dataset presented by Wei et al.~\cite{xinwei_visual_captions_paper} based on patents collected from the USPTO.
%
In this dataset, meta information including image perspectives has been automatically extracted from captions.
%
%
We processed the data to extract the most common~(more than 1000 samples) perspective labels~(e.g., \textit{left view}, \textit{perspective}) and identified a class taxonomy~(Table~\ref{tab:datasetDistribution}, right) covering 2, 4, and 7 classes. %
We utilize this information to compile the \textit{USPTO-PIP} dataset for patent image perspective~(PIP) classification from images. 
%
%


\paragraph{Evaluation Metric:}
\label{sec:evaluationmetric}
According to related work on image classification~\cite{vit_models,coca,vit_14,vit_e}, we use top-1 accuracy as a metric for evaluation. To account for the different number of test samples for image types, we compute the macro-average. 

\paragraph{Implementation Details:}
Models are trained for 200 epochs with batch size~$32$ using the respective training data for a given task~(Table~\ref{tab:datasetDistribution}) and Adaptive Moment Estimation~(Adam)~\cite{adam} with a learning rate of~$10^{-3}$. We choose the best model according to the loss of the validation data for evaluation. 

\subsection{Results for Visualization Type Classification}
\label{sec:results_image_type}
\begin{figure}[t]
  \centering
  \includegraphics[width=1.0\linewidth]{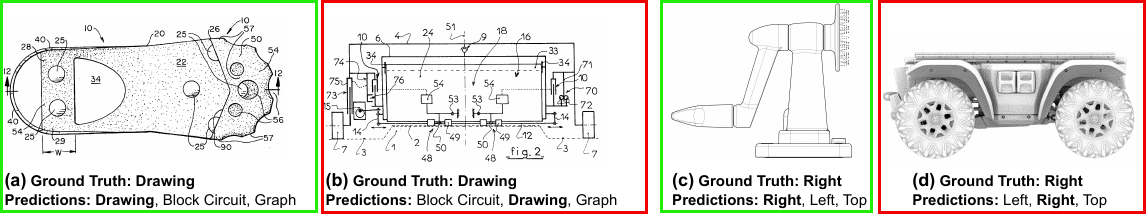}
  \caption{Examples for correct~(green) and wrong~(red) top-1  predictions.}
  \label{fig:imgTypeQuality}
\end{figure}

%
\begin{table}[t] 
  \begin{center}
    \caption{Performance of different models on the extended CLEF-IP dataset for image type classification~(left) as well as the USPTO-PIP dataset for perspective classification on different granularities~(right) with two~($C_2$), four~($C_4$), and seven classes~($C_7$).}
    \label{tab:combinedtable}
    
    \parbox{.47\linewidth}{
        \begin{tabular}{|l|c|}
          \multicolumn{2}{c}{}\\
          \hline
          \textbf{Model} & \textbf{Accuracy \%} \\
          \hline
          ResNet~\cite{he_resnet_paper} & 81.60 \\
          \hline
          EfficientNetV2~\cite{efficientnet_mingxing}  & 83.61 \\
          \hline
          \textbf{ResNeXt}~\cite{resnext_saining} & \textbf{85.01} \\
          \hline
          RegNet~\cite{regnet_ilija} & 80.20 \\
          \hline
          \hline
          CLIP~\cite{alec_clip_paper} + MLP & 82.44 \\
          \hline
          
        \end{tabular}
    }
    \hfill
    \parbox{.47\linewidth}{
        \begin{tabular}{|l|c|c|c|}
        \hline
          \textbf{Model} & \multicolumn{3}{c|}{\textbf{Accuracy \%}}  \\
          \cline{2-4}
          & \textbf{$C_2$} & \textbf{$C_4$} & \textbf{$C_7$} \\
          \hline
          ResNet~\cite{he_resnet_paper}   & 88.80 & 58.20  & 36.91   \\
          \hline
          EfficientNetV2~\cite{efficientnet_mingxing}  & 90.92 & 66.90  & 41.01 \\
          \hline
          \textbf{ResNeXt}~\cite{resnext_saining}   & \textbf{92.71} & \textbf{68.30} & \textbf{42.88} \\
          \hline
          RegNet~\cite{regnet_ilija} & 87.70 & 62.50  & 34.80\\
          \hline
          \hline
          CLIP~\cite{alec_clip_paper} + MLP & 87.15 & 59.75 & 33.40\\
          \hline
        \end{tabular}
    }
  \end{center}
\end{table}
\begin{figure}[t]
  \centering
  \includegraphics[width=\linewidth]{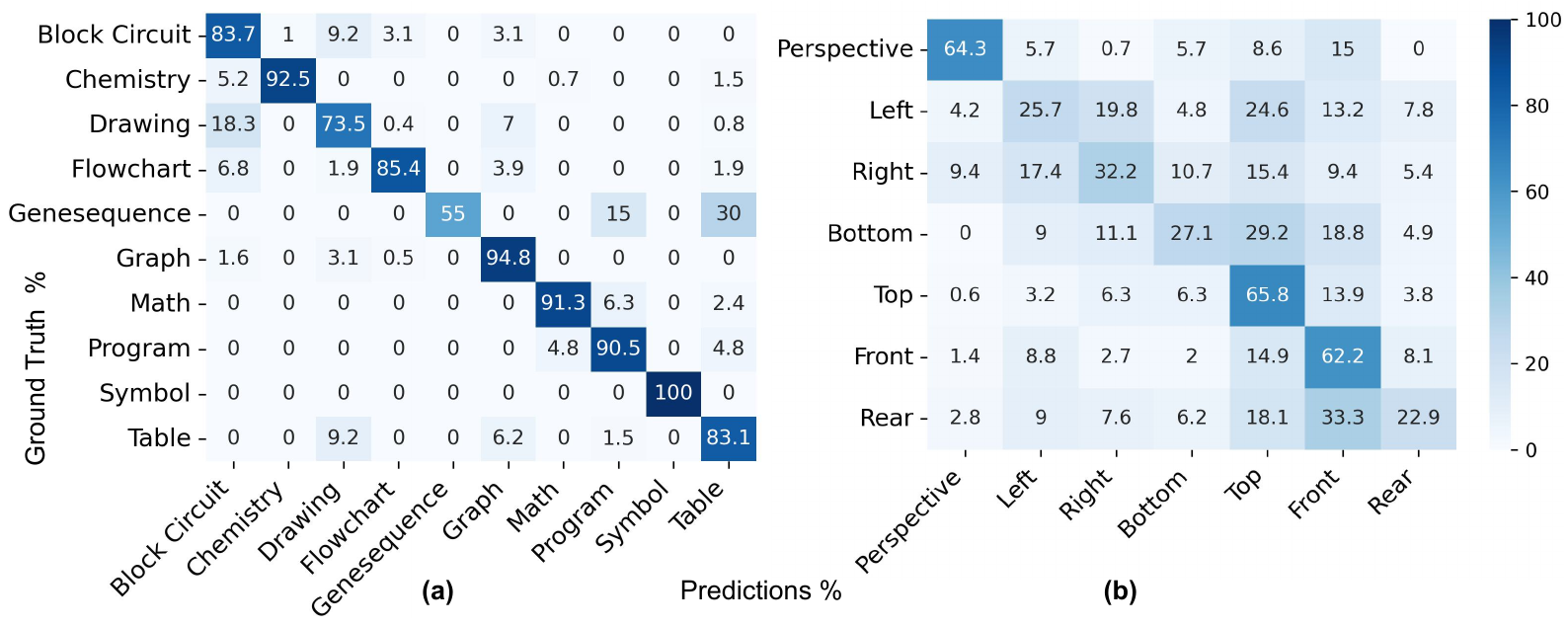}
  \caption{Confusion matrices~[\%] for patent image type~(a) and perspective classification for seven classes~($C_7$, b) using the \textit{ResNeXt} model~\cite{resnext_saining}.}
  \label{fig:combinedCF}
\end{figure}
We compared the models presented in Section~\ref{sec:methodology} on the test data of the \textit{Extended CLEF-IP 2011 Dataset}. 
According to Table~\ref{tab:combinedtable}~(left), finetuned CNNs like \textit{ResNeXt}~\cite{resnext_saining} and \textit{Efficient\-NetV2}~\cite{efficientnet_mingxing} are superior to \textit{CLIP}~\cite{alec_clip_paper}\textit{+MLP} which only finetunes an MLP for classification. However, \textit{CLIP}~\cite{alec_clip_paper}\textit{+MLP} can be finetuned much faster and with little resources while outperforming two CNN models~(\textit{ResNet}~\cite{he_resnet_paper}, \textit{RegNet}~\cite{regnet_ilija}). 
Overall, the \textit{ResNeXt} model achieves the highest accuracy~(85\%).
The confusion matrix in Figure~\ref{fig:combinedCF}(a) provides an overview of its performance for the individual image types. 
Wrong classification results can mainly be explained by the visual similarities of examples between two classes. For example, as shown in Figure~\ref{fig:imgTypeQuality}(a) and~(b), \textit{drawings} are mostly confused with \textit{block and circuit diagrams} that share similar visual elements. However, the correct prediction in these cases is within the top-2 predictions.

\subsection{Results for Perspective Classification}
\label{sec:results_perspective_type}
Results for patent image perspective classification are reported in Table \ref{tab:combinedtable}~(right). As we mentioned earlier, we consider three granularity levels with like two~($C_2$), four~($C_4$), and seven classes~($C_7$). Overall, the same conclusion can be drawn for perspective type classification~(Section~\ref{sec:results_image_type}).
%
%
%
Again, the best results are achieved by \textit{ResNeXt}~\cite{resnext_saining} for all three granularity levels.
The confusion matrix in Figure \ref{fig:combinedCF}(b) shows that most mistakes are made for classes that belong to the same parent class~(e.g., \textit{left} vs. \textit{right} side views) and thus are expected to be more similar, as also illustrated in Figure~\ref{fig:imgTypeQuality}(c) and~(d). Comparing both tasks, image perspective classification on the finest granularity is much more challenging than image type classification due to the high similarity of classes. 

\section{Conclusion}\label{sec:conclusion}
In this paper, we presented approaches based on recent deep learning models including CNNs and vision-language transformers for the classification of visualization types and perspectives in patents.
%
For this purpose, we first processed and extended available datasets from the related work for training and evaluation. 
%
%
%
In our experiments, we achieved promising results in particular using CNN-based models that outperform transformers with fixed weights in the image encoder for both tasks.
We specifically observed problems distinguishing classes with similar visual attributes. Particularly, the classification of similar image perspectives~(e.g., \textit{left} and \textit{right} side view) is very challenging.
%
%
For future work, we aim to explore more efficient finetuning techniques  for vision-language models such as prompt learning~\cite{Zhou2022ConditionalPL,DBLP:journals/ijcv/ZhouYLL22,chen2023plot} or parameter-efficient finetuning~\cite{Sung_2022_CVPR,DBLP:conf/icml/HoulsbyGJMLGAG19,DBLP:conf/nips/PanLZS022} of the entire network. Moreover, it is worth investigating hierarchy-aware models or multi-head classifiers to leverage the taxonomy of image perspectives. 
\section*{Acknowledgment}
Part of this work is financially supported by the BMBF~(Federal Ministry of Education and Research, Germany) project "ExpResViP"~(project no: 01IO2004A). We also like to thank our colleague Sushil Awale~(TIB) for his valuable feedback. 

%
%
%
\bibliographystyle{splncs04}
\bibliography{references}

\end{document}